\documentclass[runningheads]{llncs}
\usepackage{graphicx}

\usepackage{times}
\usepackage{soul}
\usepackage{url}
\usepackage[utf8]{inputenc}
\usepackage{caption}
\usepackage{amsmath}

\usepackage{booktabs}
\usepackage{algorithm}
\usepackage[noend]{algorithmic}
\usepackage{subcaption}

\usepackage{amssymb}
\usepackage{amsfonts}
\usepackage{bm}

\usepackage{amsthm}

\usepackage[misc]{ifsym}
\newtheorem*{definition*}{Definition}

\usepackage{color}

\newcommand{\bmC}{\bm{C}}
\newcommand{\bmH}{\bm{H}}
\newcommand{\bmX}{\bm{X}}
\newcommand{\bmS}{\bm{S}}
\newcommand{\bmD}{\bm{D}}
\newcommand{\bmA}{\bm{A}}
\newcommand{\bmI}{\bm{I}}
\newcommand{\bmW}{\bm{W}}
\newcommand{\bmY}{\bm{Y}}
\newcommand{\bmZ}{\bm{Z}}
\newcommand{\bmP}{\bm{P}}

\newcommand{\typical}{non-scalable}
\newcommand{\Typical}{Non-scalable}

\begin{document}
\title{GNN Transformation Framework for Improving Efficiency and Scalability}
\titlerunning{GNN Transformation Framework for Improving Efficiency and Scalability}
%
%
\author{Seiji Maekawa\inst{1} \Letter \and
Yuya Sasaki \inst{1} \and
George Fletcher \inst{2} \and
Makoto Onizuka\inst{1}}

\authorrunning{S. Maekawa et al.}
%
\institute{Osaka University, 1--5 Yamadaoka, Suita, Osaka, Japan\\
\email{\{maekawa.seiji,sasaki,onizuka\}@ist.osaka-u.ac.jp}\\
\and
Eindhoven University of Technology, P.O. Box 513, 5600 MB, Eindhoven, Netherlands\\
\email{g.h.l.fletcher@tue.nl}
}
\maketitle              
\begin{abstract}
    We propose a framework that automatically transforms \typical\ GNNs into precomputation-based GNNs which are efficient and scalable for large-scale graphs.
    The advantages of our framework are two-fold; 
    1) it transforms various \typical\ GNNs to scale well to large-scale graphs by separating local feature aggregation from weight learning in their graph convolution,
    2) it efficiently executes precomputation on GPU for large-scale graphs by decomposing their edges into small disjoint and balanced sets. 
    Through extensive experiments with large-scale graphs, we demonstrate that the transformed GNNs run faster in training time than existing GNNs while achieving competitive accuracy to the state-of-the-art GNNs. 
    Consequently, our transformation framework provides simple and efficient baselines for future research on scalable GNNs. 
    \keywords{Graph neural networks  \and Large-scale graphs \and Classification.}
\end{abstract}

\section{Introduction}
\label{sec:intro}

Graph is a ubiquitous structure that occurs in many domains, such as Web and social networks. 
As a powerful approach for analyzing graphs, Graph Neural Networks (GNNs) have gained wide research interest \cite{zhang2020deep,wu2020comprehensive}. 
Many GNNs have been proposed for node classification and representation learning including GCN~\cite{kipf2017semi}, which is the most popular GNN variant.
Most existing GNNs adopt graph convolution that performs three tasks; 1) feature aggregation, 2) learnable weight multiplication, and 3) activation function application (e.g., ReLU, a non-linear function). 
By stacking multiple graph convolutional layers, they propagate node features over the given graph topology. 
However, these existing GNNs cannot be efficiently trained on large-scale graphs since the GNNs need to perform three tasks in graph convolution every time learnable weights are updated. 
In addition, large-scale graphs cannot be put on GPU memory for efficient matrix operations. 
As a result, graph convolution is not efficient and scalable for large-scale graphs.

A major approach to apply GNNs to large-scale graphs is to separate feature aggregation from graph convolution so that GNNs can precompute aggregated features ~\cite{wu2019simplifying,sign_icml_grl2020,maurya2021improving}.
These methods are called {\it precomputation-based} GNNs.
In detail, they remove non-linearity, i.e., activation functions, from graph convolution so that feature aggregation is separated from weight learning. 
Thanks to the independence of feature aggregation and weight learning, precomputation-based GNNs are efficient in learning steps by precomputing feature aggregation before training learnable weights. 

Though some existing works tackle the scalability problem of GNNs as discussed above, most widely studied GNNs are not scalable to large-scale graphs for the following two reasons. 
First, existing studies on precomputation-based GNNs~\cite{wu2019simplifying,sign_icml_grl2020,maurya2021improving} focus on introducing several specific GNN architectures that are manually designed.
So, it is laborsome to apply the same precomputation idea to other GNNs.
An interesting observation is that they share the common motivation: precomputation of feature aggregation is indispensable for high scalability. 
To our best knowledge, there are no works that study a general framework that transforms \typical\ GNNs to scalable precomputation-based GNNs.
Second, existing precomputation schemes are not scalable because they need to put complete graphs (e.g., graphs with one billion edges~\cite{hu2020ogb}) on GPU memory. 
Since the size of large graphs typically exceeds the memory size of general GPU, existing works precompute feature aggregation on CPU. 

To tackle the above issues, we address two research questions: \textbf{Q1}: \textit{Can we design a general procedure that transforms \typical\ GNNs to efficient and scalable precomputation-based GNNs while keeping their classification performance?} and \textbf{Q2}: \textit{Can we efficiently execute the precomputation on GPU?}
There are two technical challenges which must be overcome to answer our questions. 
First, we need to automatically transform \typical\ GNNs to precomputation-based GNNs.
We should develop a common transformation procedure that can be applied to various \typical\ GNNs while preserving their expressive power. 
Second, we need to decompose large graphs into small groups each of which can be handled efficiently with GPU. 
Typically, graph decomposition suffers from an imbalance problem since node degree distributions usually follow power law distributions~\cite{newman2010intro}. 
Hence, we should divide graphs into balanced groups and select an appropriate group size so that precomputation time is optimized. 

In this paper, we propose a framework\footnote{Our codebase is available on (\url{https://github.com/seijimaekawa/LCtransformation}).} that automatically transforms \typical\ GNNs into precomputation-based GNNs with a scalable precomputation schema. 
As for the first challenge, we develop a new transformation procedure, called Linear Convolution (LC) transformation, which can be applied to various \typical\ GNNs so that transformed GNNs work efficiently and scale well to large-scale graphs. 
Our transformation procedure removes non-linear functions from graph convolution, 
but incorporates non-linear functions into weight learning.
This idea is derived from our hypothesis that it is not crucial to incorporate non-linearity into graph convolutional layers but into weight learning for prediction.
Since our transformation preserves the major functionality of graph convolution and a similar expressive power to original GNNs, the transformed GNNs can achieve competitive prediction performance to the original ones while improving their scalability. 
As for the second challenge, we develop a block-wise precomputation scheme which optimally decomposes large-scale graphs into small and balanced blocks each of which can fit into GPU memory.
We introduce a simple decomposition approach to ensure that blocks are balanced and give minimization formulas that decide the optimal block size under limited GPU memory. 

Through extensive experiments, we validate that our transformation procedure and optimized block-wise precomputation scheme are quite effective. 
First, we show that our LC transformation procedure transforms \typical\ GNNs to efficient and scalable precomputation-based GNNs while keeping their node classification accuracy. 
Second, we show that our precomputation scheme is more efficient than that of existing precomputation-based GNNs.
In summary, our transformation procedure provides simple and efficient baselines for future research on scalable GNNs by shining a spotlight on existing non-scalable methods.

The rest of this paper is organized as follows.
We describe notations and fundamental techniques for our method in Section \ref{sec:preliminaries}.
Section \ref{sec:proposal} proposes our framework. 
We give the purpose and results of experiments in Section \ref{sec:experiments}.
Section \ref{sec:related} describes the details of related work. 
Finally, we conclude this paper in Section \ref{sec:conclusion}.

\section{Preliminaries}
\label{sec:preliminaries}
An {\em undirected attributed graph with class labels} is a triple $G=(\bmA,\bmX,\bmC)$ where $\bmA \in \{0,1\}^{n\times n}$ is an adjacency matrix, $\bmX\in \mathbb{R}^{n\times d}$ is an attribute matrix assigning attributes to nodes, 
and a class matrix $\bmC\in\{0,1\}^{n\times y}$ contains class information of each node,
and $n,d,y$ are the numbers of nodes, attributes and classes, respectively. If there is an edge between nodes $i$ and $j$, $\bmA_{ij}$ and $\bmA_{ji}$ are set to one. 
We define the degree matrix $\bmD=\text{diag}(D_1,\dots,D_n)\in\mathbb{R}^{n\times n}$ as a diagonal matrix, where $D_i$ expresses the degree of node $i$.
We also define an identity matrix $\bmI=\text{diag}(1,\dots,1)\in\mathbb{R}^{n\times n}$ and an adjacency matrix extended with self-loops $\Tilde{\bmA}=\bmA + \bmI$. 
We define node embeddings $\bmH\in\mathbb{R}^{n\times h}$, where $h$ is the dimension of a hidden layer. 
We summarize notation and their definitions in Table~\ref{tb:02_table_notation}. 

\begin{table}[t]
    \centering
    \caption{Notation and definitions}
    \label{tb:02_table_notation}
\begin{tabular}{cc}\toprule
    $n$ & number of nodes\\
    $d$ & dimension of features\\
    $y$ & number of classes\\
    $h$ & dimension of hidden layer\\
    $K$ & number of hidden layers\\
    $\bmA \in \mathbb{R}^{n\times n}$ & adjacency matrix \\
    $\tilde{\bmA} \in \mathbb{R}^{N\times N}$ & extended adjacency matrix \\
    $\bmS \in \mathbb{R}^{n\times n}$ & normalized adjacency matrix \\
    $\bmX \in \mathbb{R}^{n\times d}$ & feature matrix \\
    $\bmC \in \mathbb{R}^{n\times y}$ & class matrix \\
    $\bmD \in \mathbb{R}^{n\times n}$ & degree matrix \\
    
    $\bmH \in \mathbb{R}^{n\times h}$ & node embeddings\\
    $\bmW_1 \in \mathbb{R}^{d\times h}, \bmW_2,\dots,\bmW_{K-1} \in \mathbb{R}^{h\times h}, \bmW_K \in \mathbb{R}^{h\times y}$ & weight matrices\\
    $\bmY \in \mathbb{R}^{n\times y}$ & predicted label matrix\\ 
    \bottomrule

\end{tabular}
\end{table}

\subsection{Graph Convolutional Networks}
Multi-layer GCN is a standard GCN model which was proposed in \cite{kipf2017semi}. 
GCNs learn a feature representation for the feature of each node over layers. 
For the $k$-th graph convolutional layer, we denote the input node representations of all nodes by the matrix $\bmH^{(k-1)}$ and the output node representations by $\bmH^{(k)}$. 
The initial node representations are set to the input features, i.e., $\bmH^{(0)}=\bmX$.
Let $\bmS$ denote the normalized adjacency matrix 
\begin{align}
    \label{eq:normalization}
    \bmS = \Tilde{\bmD}^{-\frac{1}{2}}\Tilde{\bmA}\Tilde{\bmD}^{-\frac{1}{2}}.
\end{align}
This normalized adjacency matrix is commonly used as a graph filter for graph convolution. 
The graph filter is known as a low-pass filter that filters out noise in node features~\cite{kipf2017semi}. 
For each layer, GCN propagates the embedding of a node to its neighbors as follows:
\begin{align}
    \bmH^{(k)} = \sigma(\bmS\bmH^{(k-1)}\bmW_k),
\end{align}
where $\bmW_k$ denotes the weight matrix of the $k$-th layer and $\sigma$ denotes a non-linear function, e.g., ReLU. 
In the output layer, $K$-layer GCN outputs a predicted label matrix $\bmY\!\in\!\mathbb{R}^{n\times y}$ as:
{
\abovedisplayskip=2mm
\belowdisplayskip=2mm
\begin{align}
    \bmY = \text{softmax}(\bmS\bmH^{(K-1)}\bmW_K),
\end{align}
}\!\!
where softmax$(\bmP)_{ij}=\frac{\exp(\bmP_{ij})}{\sum^y_{j=1} \exp(\bmP_{ij})}$ for a matrix $\bmP$.
The number of layers is typically set to $K=2$ \cite{kipf2017semi}.

\subsection{Precomputation-based GNNs}
Several precomputation-based GNNs have been proposed recently~\cite{wu2019simplifying,sign_icml_grl2020,maurya2021improving}.
Their fundamental and common idea is to remove non-linear functions between each layer in order to precompute feature aggregation.
We explain Simplifying Graph Convolution (SGC for short)~\cite{wu2019simplifying} which is the simplest precomputation-based GNN.
Thanks to the removal, $K$-layer GCN can be rewritten as follows by unfolding the recursive structure:
{
\abovedisplayskip=2mm
\belowdisplayskip=2mm
\begin{align}
    \bmY = \text{softmax}(\bmS\dots\bmS\bmX\bmW_1\ldots\bmW_K).
\end{align}
}\!\!
The repeated multiplication with the normalized adjacency matrix $\bmS$ can be simplified into a $K$-th power matrix $\bmS^K$ and the multiple weight matrices can be reparameterized into a single matrix $\bmW=\bmW_1\dots\bmW_K$. 
The output becomes
{
\abovedisplayskip=2.mm
\belowdisplayskip=2.mm
\begin{align}
\label{eq:sgc_k-th_power}
    \bmY = \text{softmax}(\bmS^K\bmX\bmW).
\end{align}
}\!\!
By separating graph feature aggregation and weight learning, SGC precomputes $\bmS^K\bmX$ before learning $\bmW$.
The other methods also follow the same idea: separating feature aggregation and weight learning and precomputing feature aggregation.

\begin{figure}[!t]
\centering
  \includegraphics[width=12.cm]{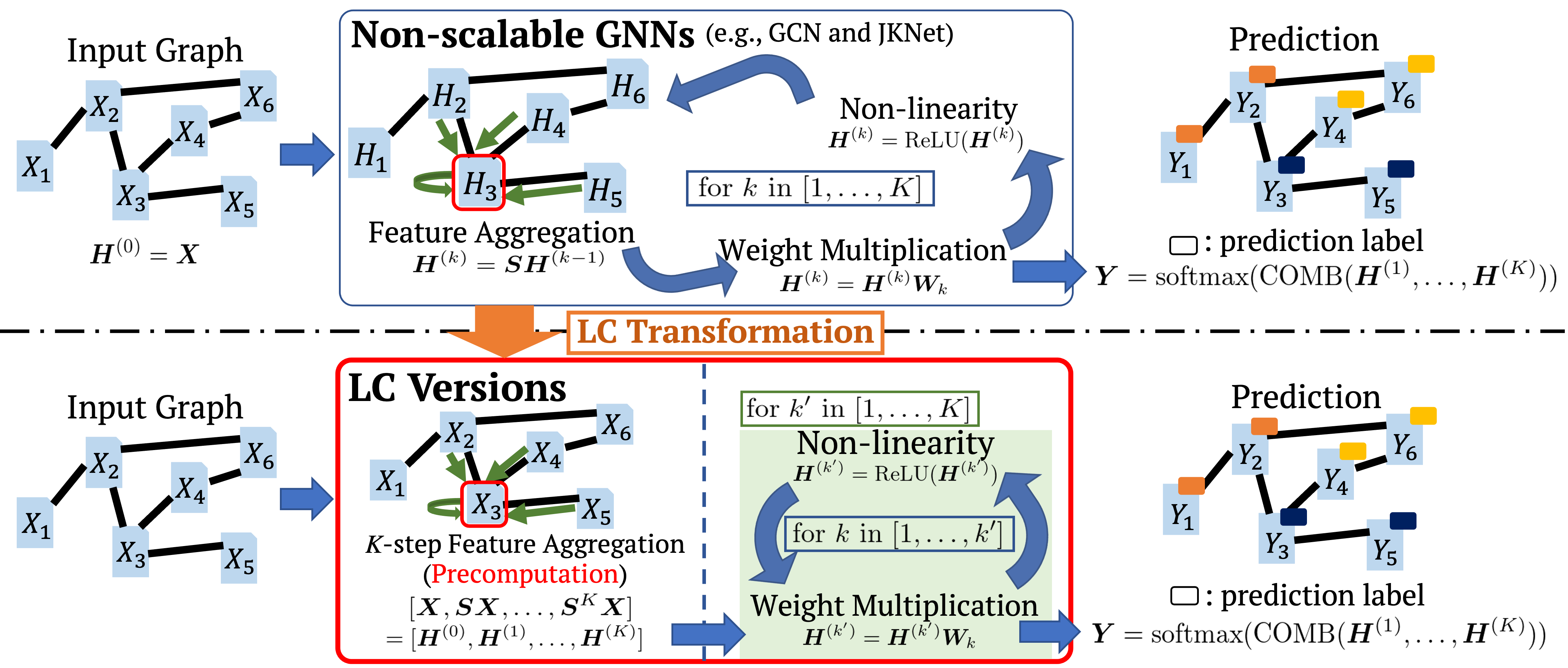}
  \vspace{-2.mm}
  \caption{Example of LC transformation. Upper part: \typical\ GNNs operate $K$-layer graph convolution combining feature aggregation, weight multiplication, and activation function application (ReLU). This example corresponds to $K$-layer GCN if $\text{COMB}$ outputs only $\bmH^K$. Lower part: LC transformation separates feature aggregation and weight learning while keeping the similar architectures with the original GNNs. LC versions avoid recomputing feature aggregation whenever learnable weights are updated at each learning step. }
  \label{fg:03_framework}
  \vspace{-2.mm}
\end{figure}
\section{GNN Transformation Framework}
\label{sec:proposal}
We propose a general framework that automatically transforms \typical\ GNNs to efficient and scalable precomputation-based GNNs and efficiently executes precomputation of feature aggregation on GPU. 
We first introduce a transformation procedure that automatically rewrites the formulations of \typical\ GNNs so that the transformed GNNs run efficiently and scale well to large-scale graphs (Section~\ref{ssec:LC_trans}). 
We also describe a limitation of our transformation, namely, that it does not support GNNs that require dynamical changes of graph filters during weight learning.
Our transformation procedure is applicable not only to GCN~\cite{kipf2017semi} but also to the state-of-the-art GNNs, such as JKNet~\cite{xu2018jumping}, H2GCN~\cite{zhu2020beyond} and GPRGNN~\cite{chien2021adaptive}. 
Next, we introduce a block-wise precomputation scheme that efficiently computes feature aggregation for large-scale graphs (Section~\ref{ssec:efficient_precomp}). 
The core idea is to decompose an adjacency matrix and feature matrix into disjoint and balanced blocks each of which can be handled on GPU. 
Also, we formulate and solve an optimization problem that decides the optimal size of blocks.
Note that this scheme is a general approach since it can be applied to existing precomputation-based GNNs ~\cite{wu2019simplifying,sign_icml_grl2020,maurya2021improving}.

\subsection{Linear Convolution Transformation} 
\label{ssec:LC_trans}
LC transformation is the first concrete procedure that transforms \typical\ GNNs to efficient and scalable precomputation-based GNNs, which have a similar functionality to the input GNNs. 
We call the output the \textit{LC version} of the input GNN. 
LC transformation is motivated by the effectiveness of SGC and Multi-Layer Perceptron (MLP).
SGC preserves the major benefit of graph convolution with efficient training by precomputing feature aggregation, but it degrades the accuracy due to the lack of non-linearity~\cite{chien2021adaptive}.
Beside, MLP outperforms linear regression in classification task by using non-linear functions but does not capture the structures of graphs.
LC version of GNN leverages both the strengths of SGC and MLP by precomputing feature aggregation and then learning weights with non-linearity.

Figure~\ref{fg:03_framework} demonstrates an example of LC transformation by comparing it with \typical\ GNNs. 
Intuitively, LC transformation separates feature aggregation from graph convolution that performs 1) feature aggregation, 2) weight multiplication, and 3) activation function application (e.g., ReLU, a non-linear function). 
Notice that a normalized adjacency matrix $\bmS$ is adjacent to the feature matrix $\bmX$ in the formulation of LC versions (see the left part of the red box of the figure). 
So, we can precompute $\bmS^k\bmX$ in the same way as SGC~\cite{wu2019simplifying}. 
Thanks to the separation, LC versions can avoid computing feature aggregation whenever learnable weights are updated at each learning step (see the right part of the red box of the figure). 
Hence, LC versions efficiently work and scale well to large-scale graphs.

\smallskip\noindent
\textbf{Discussion.}
We discuss why LC versions work from two aspects, feature aggregation and weight learning. 
As in the discussion on the spectral analysis \cite{wu2019simplifying}, feature aggregation acts as a low-pass filter that produces smooth features over the graph, which is the major benefit of graph convolution. 
In this sense, LC versions are expected to have the same functionality as the input GNNs since LC transformation preserves feature aggregation within multi-hops. 
As for weight learning, 
LC versions have a similar learning capability to their original GNNs since they have a similar model architecture of multi-layer neural networks. 
As a result, LC versions can achieve a similar prediction performance to their original GNNs while scaling to large-scale graphs.

\smallskip\noindent
\textbf{Procedure.} 
Next, we describe the procedure of LC transformation, which removes non-linear functions from graph convolution, but incorporates non-linear functions into weight learning. 
We first give the definition of LC transformation below:
\begin{definition*}[LC transformation]
    Given a \typical\ GNN algorithm, LC transformation iteratively applies a function $f_{LC}$ to the formulation of the input GNN since \typical\ GNNs have multiple graph convolutional layers.
    $f_{LC}$ commutes matrix multiplication of $\bmS$ and a non-linear function $\sigma$ as follows:
    \begin{align}
        \textstyle f_{LC}: g_2(\bmS\sigma(g_1(\bmX))) \xrightarrow[f_{LC}]{}  g_2(\sigma(\bmS g_1(\bmX))),
    \end{align}
    where $g_1$ and $g_2$ indicate any functions that input and output matrices. 
    The iteration continues until the formulation does not change. 
    LC transformation outputs a precomputation-based GNN having the transformed formulation, i.e., the LC version of the input GNN. 
\end{definition*}
To intuitively explain the details, we use JKNet \cite{xu2018jumping} as an example, which is a widely used GNN. 
The formulation of JKNet (GCN-based) is as follows: 
\begin{align}
\label{eq:jknet}
    \bmH = \text{COMB}_{k=1}^K(\bmS\sigma(\bmS\sigma(\dots(\bmS\bmX\bmW_1)\dots)\bmW_{k-1})\bmW_k), 
\end{align}
where COMB expresses a skip connection between 
different layers, such as concatenation of intermediate representations or max pooling. 
By applying a softmax function to feature representations $\bmH$, JKNet outputs a prediction result $\bmY$, i.e., $\bmY = \text{softmax}(\bmH)$. 
We apply $f_{LC}$ to it in order to transform the formulation of an input GNN. 
To this end, we assign $g_1(\bmX)=\bmS\sigma(\dots(\bmS\bmX\bmW_1)\dots)\bmW_{k-1}$ and $g_2(\bmS\sigma(g_1(\bmX))) = \text{COMB}_{k=1}^K(\bmS\sigma(g_1(\bmX))\bmW_k)$.
By utilizing $f_{LC}, g_1$, and $g_2$, we transform Eq~\eqref{eq:jknet} as follows:
\begin{align}
    \textstyle \bmH\xrightarrow[f_{LC}]{} \text{COMB}_{k=1}^K(\sigma(\bmS^2\sigma(\dots(\bmS\bmX\bmW_1)\dots)\bmW_{k-1})\bmW_k).
\end{align}
Then, we iteratively apply $f_{LC}$ to the formulation by appropriately assigning $g_1$ and $g_2$ for each iteration.
Finally, we obtain the formulation of the LC version of the input GNN, $\bmH^{LC}$, as follows:
\begin{align}
    \bmH^{LC} = \text{COMB}_{k=1}^K(\sigma(\sigma(\dots(\bmS^k\bmX\bmW_1)\dots)\bmW_{k-1})\bmW_k).
\end{align}
Then, in the same way as the input GNN, the LC version outputs a predicted label matrix $\bmY = \text{softmax}(\bmH^{LC})$.

The LC transformation procedure is applicable not only to JKNet but also to general \typical\ GNNs including APPNP~\cite{klicpera2018predict}, MixHop~\cite{abu2019mixhop}, H2GCN~\cite{zhu2020beyond}, and GPRGNN~\cite{chien2021adaptive}. 
We give another example of applying LC transformation in Appendix \ref{ap:gprgnn}.

\smallskip\noindent
\textbf{Limitation.}
Precomputation-based GNNs can use multiple graph filters such as an exact $1$-hop away adjacency matrix and Personalized PageRank diffusion matrix~\cite{klicpera2018predict}.
Those GNNs do not dynamically control the propagation of features during weight learning, since they use constant graph filters in order to precompute feature aggregation. 
Since our framework also leverages a precomputation scheme, it cannot support those existing GNNs~\cite{velivckovic2017graph,xu2018powerful,rong2019dropedge} which dynamically sample edges or modify the importance of edges during weight learning.
For example, Dropedge~\cite{rong2019dropedge} randomly reduces a certain number of edges at each iteration. 
A possible future research direction is that we simulate random edge reduction by utilizing the deviations of feature aggregation.

\subsection{Efficient Precomputation}
\label{ssec:efficient_precomp}
Existing precomputation-based GNNs need to use CPUs to compute feature aggregations for large-scale graphs since they do not fit on GPU memory. 
This CPU computation has large cost and a deteriorating effect on efficiency.

To tackle this problem, we propose a simple yet efficient block-wise precomputation scheme and provide a formulation for optimal decomposition for our block-wise precomputation scheme. 
The core idea is to decompose the edge set of a given graph into disjoint and balanced groups, while existing approaches \cite{zheng2020distdgl} decompose the node set into groups, i.e., row/column wise decomposition. 
Our scheme is inspired by edge partitioning \cite{low2012distributed,gonzalez2012powergraph}, which aims to decompose a graph into groups having similar numbers of edges such that communication costs for graph operations are minimized in distributed environments. 
Our scheme consists of three steps. 
First, it decomposes an adjacency matrix and feature matrix into small disjoint blocks each of which can be put on GPU memory. 
Second, the scheme computes block-wise matrix operations for the disjoint blocks on GPU. 
Third, it aggregates the results of the block-wise matrix operations and obtains the whole matrix operation result.

\smallskip\noindent
\textbf{Precomputation on GPU. }
There are two matrix operations to be precomputed, adjacency matrix normalization and feature aggregation. 
First, we describe the computation of adjacency matrix normalization shown by Eq.~\eqref{eq:normalization}. 
Since an adjacency matrix is typically sparse, we utilize adjacency list $(i,j)\in\mathcal{E}$, where $\tilde{\bmA}_{ij}=1$. 
To obtain small blocks each of which can be loaded on GPU memory, 
we decompose $\mathcal{E}$ into disjoint sets that include similar numbers of edges, $\mathcal{E}^{(1)}\cup\dots\cup\mathcal{E}^{(a)}$, where $a$ is a number of sets and $\mathcal{E}^{(p)}\cap\mathcal{E}^{(q)}=\emptyset \text{ if } p\ne q$. 
Note that the sizes of the sets $\mathcal{E}^{(1)},\dots,\mathcal{E}^{(a)}$ are balanced. 
Then, we decompose $\tilde{\bmA}=\tilde{\bmA}^{(1)}+\dots+\tilde{\bmA}^{(a)}$, where $\tilde{\bmA}^{(1)},\dots,\tilde{\bmA}^{(a)}\in\mathbb{R}^{n\times n}$ and $\tilde{\bmA}^{(l)}_{ij}=1$ if $(i,j)\in\mathcal{E}^{(l)}$.
Then, we can rewrite Eq.~\eqref{eq:normalization} as follows:
\begin{align}
\textstyle \bmS = \tilde{\bmD}^{-\frac{1}{2}}\tilde{\bmA}\tilde{\bmD}^{-\frac{1}{2}} = \sum_{l=1}^a\tilde{\bmD}^{-\frac{1}{2}}\tilde{\bmA}^{(l)}\tilde{\bmD}^{-\frac{1}{2}}.
\end{align}
By appropriately selecting the number of blocks $a$, $\tilde{\bmD}^{-\frac{1}{2}}\tilde{\bmA}^{(l)}\tilde{\bmD}^{-\frac{1}{2}}$ can be executed on GPU. 
We sum the results of the block-wise matrix computations.  
This summation can be efficiently computed on CPU by disjoint union of edge lists since $\mathcal{E}^{(l)}$, i.e., $\tilde{\bmA}^{(l)}$, is disjoint each other. Since our decomposition is agnostic on nodes, the decomposed blocks can be easily balanced while row/column(node)-wise decomposition approaches suffer from an imbalance problem. 
Further discussion on Limitations follows below in this subsection.

\begin{algorithm}[t]
 \begin{algorithmic}[1]
  \REQUIRE normalized adjacency matrix $\bmS$, feature matrix $\bmX$, number of layers $K$
  \ENSURE aggregated feature list $\mathit{SX\_list}$
  \STATE $\mathit{SX\_list} = []$
  \STATE $\bmS^{(1)}, \bmS^{(2)}, \dots, \bmS^{(b)} = \text{split}(\bmS)$ \hfill $\vartriangleright$ disjoint edge sets
  \STATE $\bmX_{prev} = \bmX$
  \FOR{$k=1$ to $K$}
    \STATE $\bmX^{(1)}, \bmX^{(2)}, \dots, \bmX^{(c)} = \text{split}(\bmX_{prev})$
    \FOR{$i=1$ to $c$}
        \STATE $\bmX_{tmp} = [0]^{n\times \lceil d/c\rceil}$ \hfill $\vartriangleright$ same size to $\bmX^{(i)}$
        \FOR{$j=1$ to $b$}
            \STATE $\bmZ_{tmp} = \bmS^{(j)}\bmX^{(i)}$ \hfill $\vartriangleright$ on GPU
            \STATE $\bmX_{tmp} = \bmX_{tmp} + \bmZ_{tmp}$ \hfill $\vartriangleright$ on GPU
        \ENDFOR
    \IF{$i==1$}
        \STATE $\bmX_{conc} = \bmX_{tmp}$ \hfill $\vartriangleright$ on CPU
    \ELSE
        \STATE $\bmX_{conc} = \text{concat}(\bmX_{conc},\bmX_{tmp})$ \hfill $\vartriangleright$ on CPU
    \ENDIF
    \ENDFOR
        
    \STATE $\mathit{SX\_list}$.append($\bmX_{conc}$)
    \STATE $\bmX_{prev}=\bmX_{conc}$
  \ENDFOR
 \end{algorithmic}
 \caption{Block-wise feature aggregation. }
 \label{al: feature aggregation}
\end{algorithm} 


Next, we introduce a block-wise computation for feature aggregation on GPU. 
Algorithm~\ref{al: feature aggregation} describes the procedure of the computation. 
To obtain small blocks of a normalized adjacency matrix $\bmS$, we decompose it into $\bmS^{(1)},\dots,\bmS^{(b)}\in\mathbb{R}^{n\times n}$ where $b$ is a number of blocks (line $2$). 
Similarly to the decomposition of $\bmA$, each corresponding edge list is disjoint and includes similar numbers of edges.
Also, in order to obtain small blocks of a feature matrix $\bmX$, we decompose it into $\bmX^{(1)},\dots,\bmX^{(c)}$, where $c$ is a number of blocks (line $5$). 
Since we assume that $\bmX$ is a dense matrix, we adopt column-wise decomposition, i.e., $\bmX = \text{concat}(\bmX^{(1)},\dots,\bmX^{(c)})$. 
Then, we compute matrix multiplication $\bmS^{(j)}\bmX^{(i)}$ for each pair on GPU (line $9$). 
We aggregate $\bmS^{(j)}$ by summation (line $10$) and aggregate $\bmX_{tmp}$ by concatenation (lines $11$--$14$). 
$\bmX_{prev}$ is updated by the aggregated features $\bmX_{conc}$ (line $16$). 
We repeat this aggregation $K$ times (lines $4$--$16$). 

\smallskip\noindent 
\textbf{Optimal graph decomposition. } 
We discuss an optimal decomposition for our block-wise precomputation scheme.
We have two requirements to decompose large matrices into disjoint blocks. 
First, each matrix operation for disjoint blocks can be executed on GPU.
Second, the number of disjoint blocks is as small as possible to reduce the number of block-wise matrix operations. 
To simplify the discussion, we assume that the running time of a matrix operation on GPU is the same regardless of the matrix size.

As for the block-wise adjacency matrix normalization, we minimize a number of disjoint blocks, $a$. 
We formulate the minimization as follows:
\begin{align}
\label{eq:normalization_budget}
    \textstyle \min(a), \text{ subject to } \frac{\alpha_{\bmA} B_{\bmA}+\alpha_{\bmS} B_{\bmS}}{a}+\alpha_{\bmD}B_{\bmD} \le B_\text{GPU},
\end{align}
where $\alpha_{\bmA},\alpha_{\bmS},\alpha_{\bmD}$ indicate coefficients for executing matrix operations regarding $\bmA, \bmS,\allowbreak \bmD$, respectively, and $B_{\bmA}, B_{\bmS}, B_{\bmD}, B_\text{GPU}$ indicate the volume of an adjacency matrix, the volume of a normalized adjacency matrix, the volume of a degree matrix, and the available volume of a GPU, respectively. 
As for block-wise feature aggregation, we minimize the number of pairs of disjoint blocks, $bc$. 
We formulate the minimization as follows:
\begin{align}
\label{eq:feature_agg_budget}
    \textstyle \min_{b,c}(bc),  \text{ subject to } \frac{\beta_{\bmS} B_{\bmS}}{b} + \frac{\beta_{\bmX} B_{\bmX}}{c} \le B_\text{GPU},
\end{align}
where $\beta_{\bmS},\beta_{\bmX}$ indicate coefficients for executing matrix operations regarding $\bmS, \bmX$, respectively, and $B_{\bmX}$ indicates the volume of a feature matrix. 
Note that $\alpha_{\bmA},\alpha_{\bmS},\alpha_{\bmD}, \beta_{\bmS}$, and $\beta_{\bmX}$ depend on execution environments\footnote{In real environments, users can measure $\alpha_{\bmA},\alpha_{\bmS},\alpha_{\bmD}, \beta_{\bmS}$, and $\beta_{\bmX}$ by monitoring the memory usage on small graphs, even if users do not know the details of their own environments. }. 

Next, we discuss optimization regarding Eq.~\eqref{eq:normalization_budget} and \eqref{eq:feature_agg_budget}.
As for Eq.~\eqref{eq:normalization_budget}, it is trivial to find the minimum number of blocks $a$ since there are no other parameters. 
As for Eq.~\eqref{eq:feature_agg_budget}, an exhaustive search is applicable since the number of combinations of $b$ and $c$ (natural numbers) is not large.  
Consequently, these optimization problems can be easily solved.

\smallskip\noindent
\textbf{Limitation.}
Our precomputation scheme focuses on feature aggregation on a whole graph. 
This indicates that our scheme is not suitable for node-wise operations since it may decompose the edge set of the same node into different groups. 
However, accelerating feature aggregation on a whole graph is still crucial since many graph neural networks~\cite{wu2019simplifying,sign_icml_grl2020,maurya2021improving,kipf2017semi} adopt it.

\setcounter{table}{0}
\begin{table}[!t]
    \small
    \caption{Summary of datasets. }
    \centering
    \scalebox{1}{
    \begin{tabular}{c|cccc} \toprule
        Dataset & Nodes & Edges & Features & Classes \\\midrule 
        Flickr& $89,250$ & $899,756$ & $500$ & $7$ \\ 
        Reddit& $232,965$ & $11,606,919$ & $602$ & $41$ \\ 
        arxiv& 169,343 & $1,166,243$ & $128$ & $40$ \\
        papers100M& 111,059,956 & $1,615,685,872$ & $128$ & $172$ \\ \bottomrule
    \end{tabular}
    }
\label{tb:04_dataset}
\end{table}

\section{Experiments}
\label{sec:experiments}
We design our experiments to answer the following questions; \textbf{Q1}: Can our LC transformation improve the efficiency and scalability of GNNs? \textbf{Q2}: Can our block-wise precomputation scheme accelerate precomputation?

\smallskip\noindent
\textbf{Dataset.}
We use four commonly used datasets, \texttt{Flickr}~\cite{graphsaint-iclr20}, \texttt{Reddit}~\cite{hamilton2017inductive}, \texttt{ogbn-arxiv} (arxiv for short), and \texttt{ogbn-papers100M} (papers100M for short)~\cite{hu2020ogb}. 
Table~\ref{tb:04_dataset} provides the summary of the datasets. 
The sizes of the datasets range from $9$K nodes to $110$M.

In the Flickr dataset, nodes represent images uploaded to Flickr. 
If two images share common properties such as same geographic location, same gallery, comments by the same users, there is an edge between the nodes. 
Node features represent the $500$-dimensional bag-of-words associated with the image (node). 
As for node labels, the authors of \cite{graphsaint-iclr20} scan over $81$ tags of each image and manually merged them to $7$ classes. 
In the Reddit dataset, nodes represent posts. 
If the same user left comments on two posts, then there is an edge between the two posts.
Node features are the embedding of the contents of the posts. 
The labels of nodes indicate communities which the nodes belong to. 
In the ogbn-arxiv dataset, nodes represent ARXIV papers and edges indicate that one paper cites another one. 
Node features represent $128$-dimensional feature vectors obtained by averaging the embeddings of words in titles and abstracts. 
Node labels indicate subject areas of ARXIV CS papers\footnote{\url{https://arxiv.org/archive/cs}}. 
In the ogbn-papers100M (papers100M) dataset, its graph structure and node features are constructed in the same way as ogbn-arxiv. 
Among its nodes, approximately $1.5$ million nodes are labeled with one of ARXIV's subject areas. 
As in \cite{shadow_GNN}, Flickr and Reddit are under the inductive setting.
ogbn-arxiv and ogbn-papers100M are under the transductive setting.

\smallskip\noindent
\textbf{Baseline.}
We compare three types of existing methods as baselines; \typical\ GNNs, precomputation-based GNNs, and sampling-based GNNs which are scalable but inefficient (we discuss the details in Section~\ref{sec:related}). 
As for \typical\ GNNs, we use GCN\footnote{\url{https://github.com/tkipf/pygcn}}~\cite{kipf2017semi}, JKNet\footnote{Since official codes of JKNet from the authors are not provided, we simply implement JKNet based on the implementation of GCN.}~\cite{xu2018jumping}, and GPRGNN\footnote{\url{https://github.com/jianhao2016/GPRGNN}}~\cite{chien2021adaptive}.
As for precomputation-based GNNs, we use SGC\footnote{\url{https://github.com/Tiiiger/SGC}}~\cite{wu2019simplifying} and FSGNN\footnote{\url{https://github.com/sunilkmaurya/FSGNN}}~\cite{maurya2021improving}.
As for sampling-based GNNs, we use ShaDow-GNN\footnote{\url{https://github.com/facebookresearch/shaDow_GNN}}~\cite{shadow_GNN}.
FSGNN and ShaDow-GNN are the state-of-the-art precomputing-based and sampling-based GNNs, respectively. 
We note that we use our block-wise precomputation to the precomputation-based GNNs instead of using its original CPU computation for a fair comparison. 

\smallskip\noindent
\textbf{Setup.}
We tune hyperparameters on each dataset by Optuna~\cite{akiba2019optuna} and use Adam optimizer~\cite{kingma2014adam}. 
We adopt mini-batch training for precomputation-based GNNs, sampling-based GNNs, and LC-versions to deal with large-scale graphs\footnote{We will provide hyperparameter search space and the best parameters to reproduce experiments on our codebase that will be publicly available on acceptance.}. 
As for ShaDow-GNN, we use the best hyperparameter sets provided by the authors and adopt GAT~\cite{velivckovic2017graph} as a backbone model since ShaDow-GAT achieves the best accuracy in most cases reported in the paper. 
We measure training time on a NVIDIA Tesla V100S GPU (32GB) and Intel(R) Xeon(R) Gold 5220R CPUs (378GB).

\begin{table}[t]
    \centering
    \caption{Comparison on test accuracy. We report the average values (standard deviation) over 5 runs.}
    \scalebox{1}{
    \begin{tabular}{l|ccc}\toprule
         &  {Flickr} & {Reddit} & {arxiv} \\ \midrule 
        GCN   & $0.525 (0.003)$ & $0.945 (0.000)$ & $0.702 (0.005)$ \\
        JKNet & $0.526 (0.004)$ & $0.941 (0.006)$ & $0.712 (0.001)$ \\
        GPRGNN & $0.494 (0.006)$ & $0.918 (0.012)$ & $0.694 (0.006)$ \\ \hline
        SGC & $0.494 (0.037)$ & $0.948 (0.001)$ & $0.692 (0.004)$ \\
        FSGNN & $0.513 (0.001)$ & $0.964 (0.001)$ & $0.722 (0.003)$ \\ \hline
        ShaDow-GAT &  $0.531 (0.003)$ & $0.947 (0.003)$ & $0.716 (0.004)$ \\ \hline
        GCN\_LC & $0.515 (0.003)$ & $0.947 (0.001)$ & $0.710 (0.001)$ \\
        JKNet\_LC & $0.517 (0.004)$ & $0.951 (0.000)$ & $0.710 (0.003)$ \\
        GPRGNN\_LC & $0.513 (0.001)$ & $0.961 (0.000)$ & $0.720 (0.004)$ \\ 
        \bottomrule
    \end{tabular}
    }
    \label{tb:04_accuracy}
    \vspace{-2.mm}
\end{table}

\begin{table}[t]
    \centering
    \caption{Comparison on training time (per epoch/total). 
    Note that total training time includes precomputation time for SGC, FSGNN, ShaDow-GAT, GCN\_LC, JKNet\_LC, and GPRGNN\_LC. We report the average values over 5 runs. }
    \scalebox{0.95}{
    \begin{tabular}{l|ccc} \toprule
         &  {Flickr} & {Reddit} &  {arxiv} \\  \midrule
        GCN & $64.62$[ms] / $129.24$[s] & $654.70$[ms] / $1309.40$[s] & $210.81$[ms] / $421.63$[s] \\
        JKNet & $170.43$[ms] / $253.25$[s] & $1428.51$[ms] / $2552.45$[s] & $529.05$[ms] / $1058.10$[s] \\
        GPRGNN & $272.86$[ms] / $539.48$[s] & $1456.01$[ms] / $2806.62$[s] & $523.08$[ms] / $961.76$[s] \\ \hline
        SGC & $51.18$[ms] / $30.31$[s] & $141.68$[ms] / $285.43$[s] & $50.27$[ms] / $42.23$[s] \\
        FSGNN & $346.97$[ms] / $133.63$[s] & $1066.66$[ms] / $1793.91$[s] & $284.73$[ms] / $382.67$[s] \\ \hline
        ShaDow-GAT & $120.85$e3[ms] / $3634.65$[s] & $376.42$e3[ms] / $11321.09$[s] & $163.67$e3[ms] / $4913.29$[s]\\ \hline
        GCN\_LC &$56.75$[ms] / $49.85$[s] & $165.73$[ms] / $212.16$[s] & $62.59$[ms] / $120.60$[s] \\
        JKNet\_LC & $144.78$[ms] / $78.24$[s] & $430.41$[ms] / $865.71$[s] & $138.52$[ms] / $277.63$[s] \\
        GPRGNN\_LC & $287.54$[ms] / $164.88$[s] & $818.13$[ms] / $1645.49$[s] & $219.66$ [ms] / $204.56$[s] \\ \bottomrule
    \end{tabular}
    }    
    \label{tb:04_runtime}
    \vspace{-2.mm}
\end{table}

\subsection{Effectiveness of LC Transformation (Q1)}
Table~\ref{tb:04_accuracy} shows the test accuracy of LC versions and the baselines. 
LC versions (GCN\_LC, JKNet\_LC, and GPRGNN\_LC) achieve comparable test accuracy with their original GNNs (GCN, JKNet, and GPRGNN) for all datasets. 
Next, Table~\ref{tb:04_runtime} shows the training time of LC versions and the baselines. 
The LC versions run faster than their original GNNs.
Note that LC versions tend to stop earlier than \typical\ GNNs since LC versions train their models more times due to mini-batch training. 
For example, in Flickr data LC versions more efficiently train than \typical\ GNNs even if they have similar training time per epoch.
These results indicate that our framework transforms \typical\ GNNs to efficient precomputation-based GNNs with the comparable classification accuracy to the original GNNs.

\begin{table}[!t]
    \centering
    \caption{Results on papers100M. We show test/validation accuracy (standard deviation) and training time (per epoch / total). Total training time includes precomputation time. OOM indicates that the execution is out of memory.}
    \begin{tabular}{l|c|c|c} \toprule
         & Test accuracy & Val accuracy & Time (epoch / total)  \\ \midrule
        GCN &OOM&OOM&OOM\\
        JKNet &OOM&OOM&OOM\\
        GPRGNN &OOM&OOM&OOM\\ \hline
        SGC &  $0.623 (0.007)$  & $0.667 (0.002)$  & $425.15$[ms] / $2211.23$[s] \\
        FSGNN  &  $0.665 (0.003)$ & $0.706 (0.001)$&$3550.82$[ms] / $8612.48$[s] \\ \hline
        ShaDow-GAT & $0.666 (0.003)$&$0.703 (0.001)$&$2948.50$e3[ms] / $92264.76$[s] \\ \hline
        GCN\_LC&$0.647 (0.006)$&$0.688 (0.002)$&$611.90$[ms] / $2477.55$[s] \\
        JKNet\_LC&$0.641 (0.003)$&$0.689 (0.004)$&$1488.80$[ms] / $3396.69$[s] \\ 
        GPRGNN\_LC &$0.658(0.002)$&$0.696 (0.001)$&$2749.27$[ms] / $7410.47$[s] \\ \bottomrule
    \end{tabular}
    \label{tb:04_papers}
    \vspace{-2.mm}
\end{table}
\begin{figure}[!t]
\centering
  \includegraphics[width=9.cm]{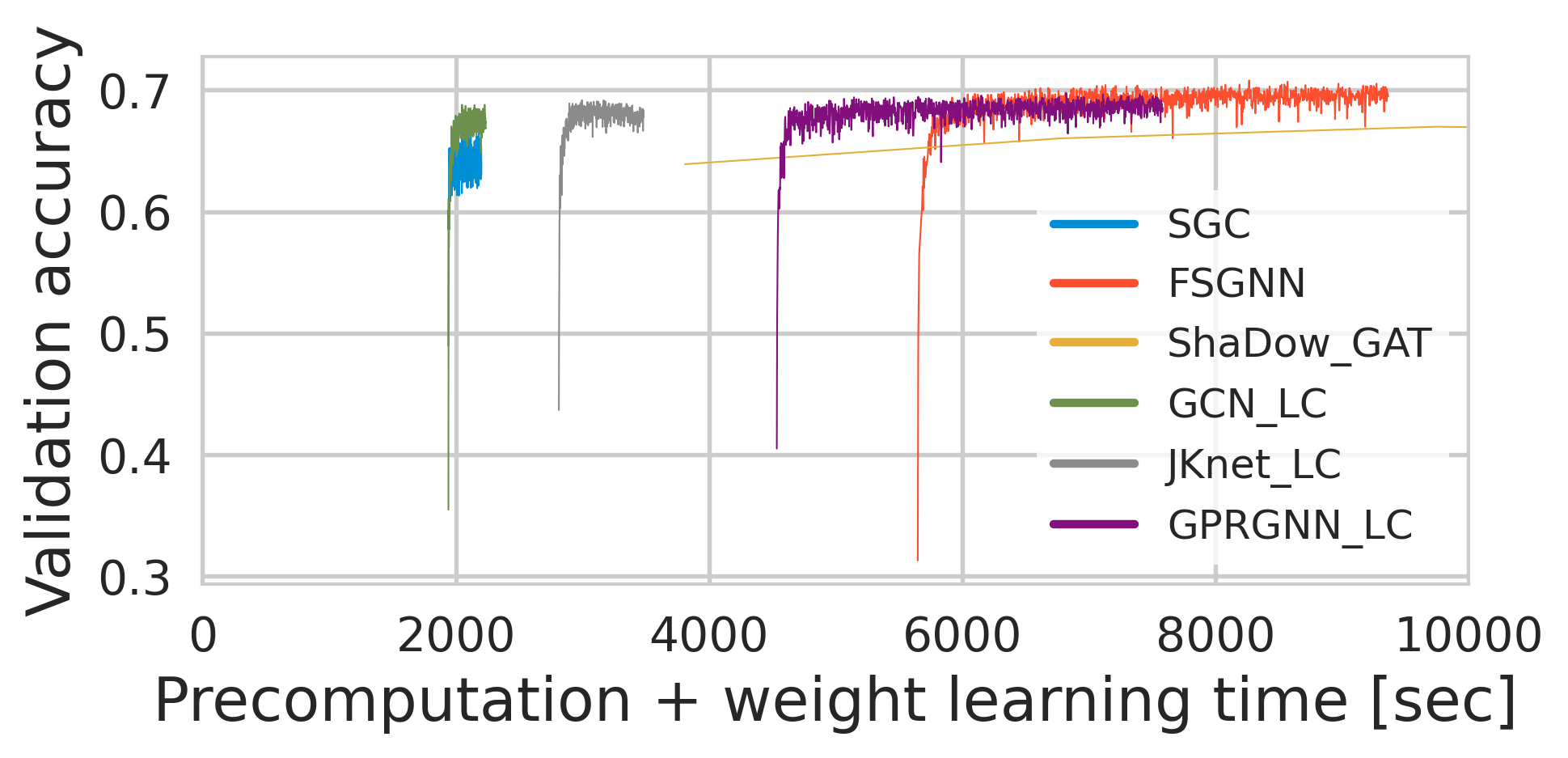}
  \vspace{-3.5mm}
  \caption{Validation accuracy over training time (precomputation and weight learning time) on papers100M. Plots indicate epochs. LC versions (GCN\_LC, JKNet\_LC, and GPRGNN\_LC) are faster than FSGNN and ShaDow-GAT while achieving competitive accuracy. }
  \label{fg:04_validation_time}
  \vspace{-2mm}
\end{figure}

\smallskip\noindent
\textbf{Comparison on large-scale graph. }
Table~\ref{tb:04_papers} shows the performance comparison on papers100M having more than $100$ million nodes and one billion edges.
\Typical\ GNNs (GCN, JKNet, and GPRGNN) cannot work on papers100M since the whole graph cannot be put on GPU memory. 
GPRGNN\_LC achieves comparable accuracy (approximate one percent difference) with FSGNN, which is the state-of-the-art precomputation-based GNN while GPRGNN\_LC runs faster than FSGNN. 
Though ShaDow-GAT achieves the highest accuracy, it requires more than $10\times$ total training time than other models. 
This is because it needs to operate graph convolutions on many enclosing subgraphs extracted from the whole graph. 
SGC obtains lower accuracy than GCN\_LC. 
This result validates that non-linearity contributes to weight learning for better classification.

In order to analyze the results on papers100M in details, we show the validation accuracy at each epoch over total training time in Figure~\ref{fg:04_validation_time}. 
Note that total training time consists of precomputation and weight learning time. 
We observe that GCN\_LC, JKNet\_LC, and GPRGNN\_LC are plotted in the upper left corner of the figure. 
This observation indicates that they require less total training time than FSGNN and ShaDow-GAT. 
The LC versions achieve competitive performance with them. 
Through these experiments, we demonstrate that LC versions are efficient and scalable for large-scale graphs. 

\subsection{Precomputation Efficiency (Q2)}
To validate the efficiency of our block-wise precomputation, we compare it with naive CPU computation adopted by existing works \cite{hu2020ogb,maurya2021improving}. 
We use a large-scale graph, papers100M, which requires a $67$GB normalized adjacency matrix and a $57$GB feature matrix. 
For adjacency matrix normalization, we set the number of disjoint blocks of an adjacency matrix to $a=3$, which satisfies Eq.~\eqref{eq:normalization_budget}. 
Also, for feature aggregation we set numbers of disjoint blocks of a normalized adjacency matrix and feature matrix to $b=10, c=16$, respectively, which satisfy Eq.~\eqref{eq:feature_agg_budget}. 

\begin{figure}[!t]
\centering
  \includegraphics[width=8.cm]{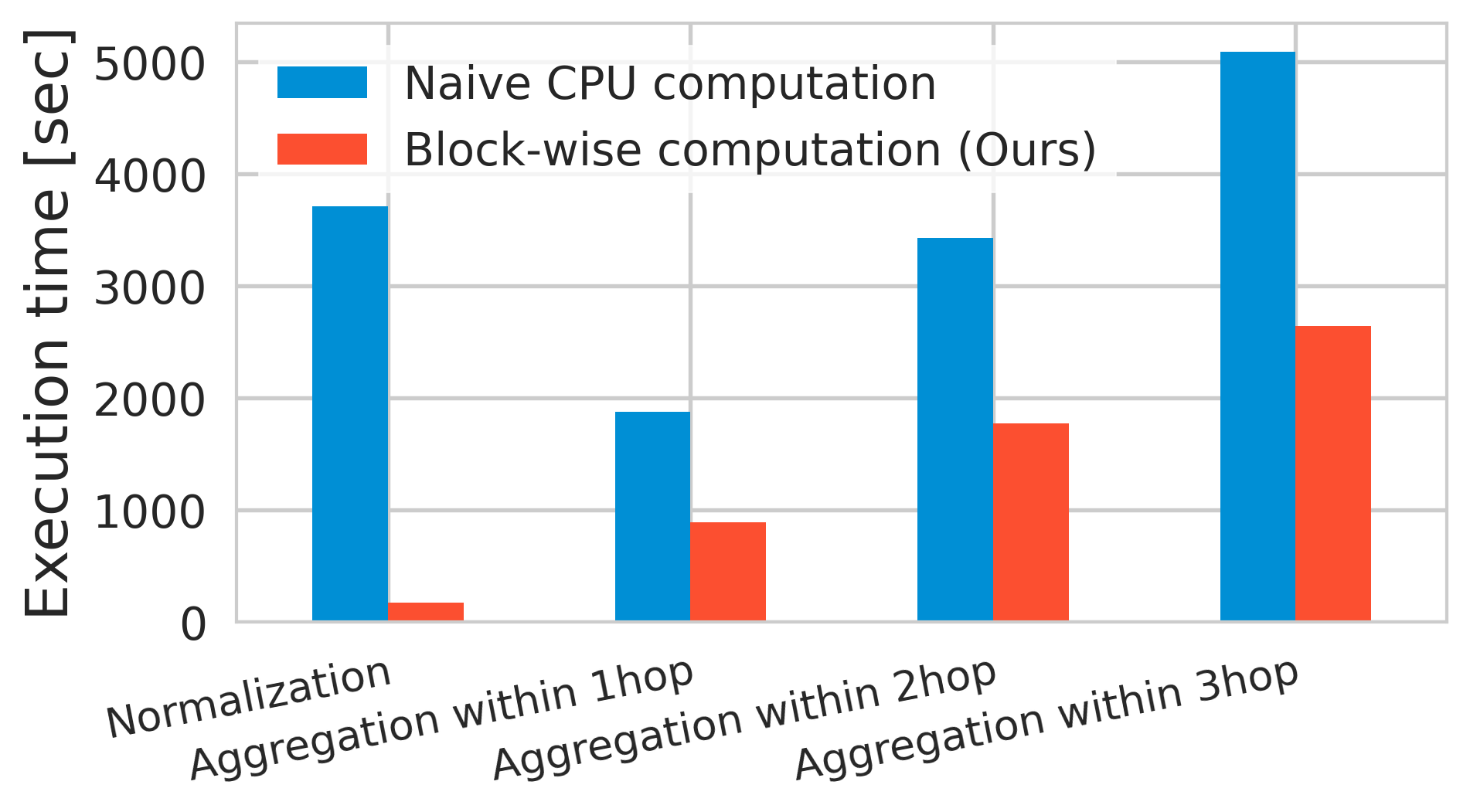}
  \vspace{-3.5mm}
  \caption{Precomputation time comparison between a naive CPU computation and our block-wise computation.}
  \label{fg:04_cpuVSgpu}
  \vspace{-2mm}
\end{figure}

Figure \ref{fg:04_cpuVSgpu} shows the precomputation time for normalization and feature aggregation on CPU and GPU. 
The result demonstrates that our block-wise precomputation is $20\times$ faster than CPU computation for normalization. 
Also, the result indicates that our precomputation is up to twice faster than CPU computation for feature aggregation. 
Hence, we conclude that our precomputation is more efficient than CPU computation on a single machine. 

\section{Related Work}
\label{sec:related}
\noindent
\textbf{Relationship between \typical\ GNNs and LC versions.}
We discuss the background of \typical\ GNNs and their LC versions. 
Graph convolution is motivated by the $1$-dim Weisfeiler-Lehman (WL-$1$) algorithm~\cite{wltest1968} which is used to test graph isomorphism; two graphs are called isomorphic if they are topologically identical. 
WL-$1$ iteratively aggregates the labels of nodes and their neighbors, and hashes the aggregated labels into unique labels.
The algorithm decides whether two graphs are isomorphic or not by using the labels of nodes at some iteration. 
\Typical\ GNNs such as GCN~\cite{kipf2017semi} replace the hash function of WL-$1$ with a graph convolutional layer which consists of feature aggregation, weight multiplication, and non-linear function application. 
As for LC versions, they replace the hash function of WL-$1$ with feature aggregation. 
These observations indicate that WL-$1$ is analogous to feature aggregation of LC versions, similarly to graph convolution of \typical\ GNNs. 

\smallskip\noindent
\textbf{Sampling-based GNNs.}
Sampling-based GNNs~\cite{hamilton2017inductive,chen2018fastgcn,chiang2019cluster,graphsaint-iclr20,shadow_GNN} avoid keeping a whole graph on GPU by computing node representations from enclosing subgraphs of the input graph. 
The major drawback of the sampling-based GNNs is that they need costly training time since they need to operate graph convolutions on many enclosing subgraphs extracted from the input graph.

\smallskip\noindent
\textbf{GNNs dynamically modifying the importance of edges.}
As we discussed in Section~\ref{ssec:LC_trans}, our transformation cannot support GNNs which dynamically control the propagation of features during weight learning. 
An example of such GNNs is GAT~\cite{velivckovic2017graph}, which learns attention parameters controlling the importance of edges for each iteration.
Another example is GIN~\cite{xu2018powerful} learns a parameter controlling a weight between self features and features from neighbors. 
One possible direction is that we first determine the parameters by training on a subset of an input graph, then fix them in order to precompute feature aggregation. 

\smallskip\noindent
\textbf{Distributed matrix operations.}
Matrix operations can be parallelized for distributed computing \cite{boehm2016systemml} \cite{awaysheh2021big}. 
For example, the authors of \cite{he2008mars} proposed Mars which is an approach for hiding the programming complexity of MapReduce on GPU. 
Also, MR-Graph \cite{qiao2015customizable} is a customizable and unified framework for GPU-based MapReduce. 
It allows its users to implement their applications more flexibly.
As for distributed graph neural network training, DistDGL~\cite{zheng2020distdgl} has proposed mini-batch training on graphs, which scales beyond a single machine. 
It suffers from an imbalance problem since it uses a typical graph clustering algorithm METIS~\cite{karypis1998fast} to partition large-scale graphs into subgraphs, while our scheme can partition an edge set into balanced subsets. 
For further scale up of graphs, it would be important to combine distributed computing and our block-wise precomputation for graphs. 

\section{Conclusion}
\label{sec:conclusion}
We presented a framework that automatically transforms \typical\ GNNs to efficient and scalable precomputation-based GNNs. 
There are two major characteristics of our framework: 1) it supports a novel transformation procedure that transforms \typical\ GNNs to efficient and scalable precomputation-based GNNs having a similar functionality to the original GNNs, 2) the precomputation of the transformed GNNs can be efficiently executed by our block-wise precomputation scheme that decomposes large-scale graphs into disjoint and balanced blocks each of which can be handled on GPU memory. 
Through our experiments, we demonstrated that the transformed GNNs run more efficiently than their original GNNs and can be scaled to graphs with millions of nodes and billions of edges. 
Due to the strong performance of LC versions, we argue that LC versions will be beneficial as baseline comparisons for future research on scalable GNNs.

\noindent\smallskip
\textbf{Acknowledgement. }
This work was supported by JSPS KAKENHI Grant Numbers JP20H00583 and JST PRESTO Grant Number JPMJPR21C5.

\bibliographystyle{splncs04}
\bibliography{reference}%

\appendix
\section{LC Version of GPRGNN}
\label{ap:gprgnn}
We show an example of LC transformation for the state-of-the-art GNN model, GPRGNN~\cite{chien2021adaptive}.
We give the formulation of GPRGNN as follows:
\begin{align}
\label{eq:GPRGNN}
    \textstyle \bmH=\sum_{k=0}^K\gamma_k\bmS^k(\sigma(\dots\sigma(\bmX\bmW_1)\dots)\bmW_T),
\end{align}
where $\gamma_k$ is an attention parameter learning the importance of $k$-th layer and $T$ is the number of layers for Multi-layer perceptrons.
Note that $S^k$ cannot be efficiently precomputed since the number of non-zero elements significantly increases when $k\ge2$ for large-scale graphs.
By iteratively applying $f_{LC}$ to Eq.~\eqref{eq:GPRGNN}, we obtain the formulation of its LC version as follows:
\begin{align}
    \textstyle \bmH^{LC}=\sum_{k=0}^K\gamma_k(\sigma(\dots\sigma(\bmS^k\bmX\bmW_1)\dots)\bmW_T).
\end{align}
$\bmS^k\bmX$ can be precomputed since it does not need to be updated when learnable weights $\bmW_1\dots\bmW_T$ and a parameter $\gamma$ are updated.

\section{Experimental Details}

\subsection{Hyperparameters}
For experiments of baselines and LC versions, we set hyperparameters similarly to \cite{maurya2021improving},
\begin{itemize}
    \item Activation: ReLU
    \item Batch size for precomputation-based GNNs: $4096$
    \item Patience: $300$
    \item Maximum number of epochs: $2000$
    \item Hidden dimension: $256$
\end{itemize}

We show hyperparameter search space for each GNN as follows. 

\smallskip\noindent
\textbf{GCN. }
The hyperparameter search space of GCN is listed as follows:
\begin{itemize}
    \item Weight decay: $[1e$-$8,1e$-$7,1e$-$6,1e$-$5]$
    \item Learning rate: $[0.005, 0.01, 0.02]$
    \item Dropout: $[0, 0.5]$
    \item Number of layers: $[2]$
\end{itemize}

\smallskip\noindent
\textbf{JKNet. }
The hyperparameter search space of JKNet is listed as follows:
\begin{itemize}
    \item Weight decay: $[1e$-$8,1e$-$7,1e$-$6,1e$-$5]$
    \item Learning rate: $[0.001, 0.005, 0.01]$
    \item Dropout: $[0, 0.5]$
    \item Number of layer: $[3]$
    \item Pooling: [concat, max]
\end{itemize}

\smallskip\noindent
\textbf{GPRGNN. }
The hyperparameter search space of GPRGNN is listed as follows:
\begin{itemize}
    \item Weight decay: $[0, 1e$-$10, 1e$-$9, 1e$-$8]$
    \item Learning rate: $[0.005, 0.01, 0.02]$
    \item Dropout for MLP: $[0.5]$
    \item Dropout before propagation: $[0,0.5,0.7]$
    \item Parameter initializing attention: $[0,0.1,0.5,0.9,1]$
    \item Number of propagation layer: $[5]$
    \item Number of layer of MLP: $[4]$
\end{itemize}

\smallskip\noindent
\textbf{SGC. }
The hyperparameter search space of SGC is listed as follows:
\begin{itemize}
    \item Weight decay: $[1e$-$9,1e$-$8,1e$-$7,1e$-$6,1e$-$5]$
    \item Learning rate: $[1e$-$4, 0.001, 0.01]$
    \item Number of layers: $[2]$
\end{itemize}

\smallskip\noindent
\textbf{FSGNN. }
The hyperparameter search space of FSGNN is listed as follows:
\begin{itemize}
    \item Weight decay1: $[1e$-$5,1e$-$4,1e$-$3]$
    \item Learning rate1: $[1e$-$4, 0.001, 0.01]$
    \item Weight decay2: $[1e$-$6,1e$-$5,1e$-$4,1e$-$3]$
    \item Learning rate2: $[1e$-$4, 0.001, 0.01]$
    \item Weight decay3: $[1e$-$6,1e$-$5,1e$-$4,1e$-$3]$
    \item Learning rate3: $[1e$-$4, 0.001, 0.01]$
    \item Weight decay for attention: $[1e$-$6,1e$-$5,1e$-$4,0.001]$
    \item Learning rate for attention: $[0.001, 0.01, 0.1]$
    \item Dropout1: $[0.5,0.6,0.7]$
    \item Dropout2: $[0.5,0.6,0.7]$
    \item Number of layers: $[3]$
\end{itemize}

\smallskip\noindent
\textbf{GCN\_LC. }
The hyperparameter search space of GCN\_LC is listed as follows:
\begin{itemize}
    \item Weight decay: $[1e$-$7,1e$-$6,1e$-$5,1e$-$4]$
    \item Learning rate: $[5e$-$5, 1e$-$4, 0.001, 0.01]$
    \item Dropout: $[0, 0.5]$
    \item Number of layers: $[2]$
\end{itemize}

\smallskip\noindent
\textbf{JKNet\_LC. }
The hyperparameter search space of JKNet\_LC is listed as follows:
\begin{itemize}
    \item Weight decay: $[1e$-$6,1e$-$5,1e$-$4, 0.001]$
    \item Learning rate: $[1e$-$4, 0.001]$
    \item Dropout: $[0, 0.5]$
    \item Number of layers: $[3]$
    \item Pooling: [concat, max]
\end{itemize}

\smallskip\noindent
\textbf{GPRGNN\_LC. }
The hyperparameter search space of GPRGNN\_LC is listed as follows:
\begin{itemize}
    \item Weight decay: $[0, 1e$-$8, 1e$-$7]$
    \item Learning rate: $[1e$-$4, 0.001, 0.01]$
    \item Dropout for MLP: $[0.5]$
    \item Dropout before propagation: $[0,0.5,0.7]$
    \item Parameter initializing attention: $[0,0.1,0.5,0.9,1]$
    \item Number of propagation layer: $[5]$
    \item Number of layer of MLP: $[4]$
\end{itemize}

We choose the best parameter set from these candidates by utilizing Optuna~\cite{akiba2019optuna} for $50$ trials. 
Configuration of baselines and LC versions to reproduce experiments in our paper is shown in Table~\ref{tb:best_param}.

\begin{table*}[t]
    \small
    \caption{Configuration for Table~\ref{tb:04_accuracy} and \ref{tb:04_papers}. OOM indicates that the execution is out of memory. ``-" indicates that the parameter is not used for the GNN. }
    \centering
    \scalebox{0.75}{
    \begin{tabular}{cc|cccccc}\toprule
         & Dataset & \begin{tabular}{c} Weight \\ decay \end{tabular} & \begin{tabular}{c} Learning\\ rate \end{tabular} & Dropout & Number of layers & Pooling & \begin{tabular}{c} Attention\\ initialing parameter \end{tabular} \\\midrule 
         & Flickr & $1$e-$6$ & $0.02$ & $0$ & $2$ & -& -\\
        GCN & Reddit & $1$e-$7$ & $0.02$ & $0$ & $2$ & -& -\\
        & arxiv & $1$e-$5$ & $0.02$ & $0$ & $2$ & -& -\\
        & papers100M & OOM & OOM & OOM & OOM & OOM & OOM \\ \hline
         & Flickr & $1$e-$5$ & $0.01$ & $0$ & $3$ & concat & -\\
        JKNet & Reddit & $1$e-$7$ & $0.005$ & $0.5$ & $3$ & concat & -\\
        & arxiv & $1$e-$7$ & $0.001$ & $0.5$ & $3$ & concat & -\\
        & papers100M & OOM & OOM & OOM & OOM & OOM & OOM \\ \hline
        & Flickr & $0$ & $0.005$ & \begin{tabular}{c}MLP : $0.5$\\ Propagation : $0.5$ \end{tabular} & \begin{tabular}{c}MLP: $4$\\ Propagation : $5$ \end{tabular} & - & $0.1$ \\
        GPRGNN & Reddit & $1$e-$9$ & $1$e-$4$ & \begin{tabular}{c}MLP : $0.5$\\ Propagation : $0.5$ \end{tabular} & \begin{tabular}{c}MLP: $4$\\ Propagation : $5$ \end{tabular} & - & $0.1$\\
        & arxiv & $1$e-$9$ & $0.005$ & \begin{tabular}{c}MLP : $0.5$\\ Propagation : $0$ \end{tabular} & \begin{tabular}{c}MLP: $4$\\ Propagation : $5$ \end{tabular} & - & $0.9$\\
        & papers100M & OOM & OOM & OOM & OOM & OOM & OOM \\ \hline
         & Flickr & $1$e-$6$ & $0.01$ & - & $2$ & - & - \\
        SGC & Reddit & $1$e-$9$ & $0.01$ & - & $2$ & - & -\\
        & arxiv & $1$e-$9$ & $0.01$ & - & $2$ & - & -\\
        & papers100M & $1$e-$8$ & $0.01$ & - & $2$ & - & - \\ \hline
        & Flickr & \begin{tabular}{c}wd1 : $1$e-$4$\\ wd2 : $1$e-$4$ \\ wd3 : $1$e-$4$ \\ wd\_att : $0.01$ \end{tabular} & \begin{tabular}{c}lr1 : $1$e-$4$\\ lr2 : $0.001$ \\ lr3 : $0.001$ \\ lr\_att : $0.001$ \end{tabular} & \begin{tabular}{c}dp1 : $0.7$\\ dp2 : $0.7$ \end{tabular} & $3$ & - & - \\
        FSGNN & Reddit & \begin{tabular}{c}wd1 : $1$e-$5$\\ wd2 : $0.001$\\ wd3 : $1$e-$4$ \\ wd\_att : $0.001$ \end{tabular} & \begin{tabular}{c}lr1 : $1$e-$4$\\ lr2 : $0.001$ \\ lr3 : $0.001$ \\ lr\_att : $0.01$ \end{tabular} & \begin{tabular}{c}dp1 : $0.6$\\ dp2 : $0.5$ \end{tabular} & $3$ & - & - \\
        & arxiv & \begin{tabular}{c}wd1 : $1$e-$5$\\ wd2 : $1$e-$5$ \\ wd3 : $1$e-$4$ \\ wd\_att : $0.001$ \end{tabular} & \begin{tabular}{c}lr1 : $1$e-$4$\\ lr2 : $0.001$ \\ lr3 : $0.001$ \\ lr\_att : $0.01$ \end{tabular} & \begin{tabular}{c}dp1 : $0.7$\\ dp2 : $0.6$ \end{tabular} & $3$ & - & - \\
        & papers100M & \begin{tabular}{c}wd1 : $1$e-$4$\\ wd2 : $1$e-$5$ \\ wd3 : $1$e-$6$ \\ wd\_att : $1$e-$4$ \end{tabular} & \begin{tabular}{c}lr1 : $1$e-$4$\\ lr2 : $1$e-$4$ \\ lr3 : $0.001$ \\ lr\_att : $0.1$ \end{tabular} & \begin{tabular}{c}dp1 : $0.5$\\ dp2 : $0.5$ \end{tabular} & $3$ & - & - \\ \hline
         & Flickr & $1$e-$5$ & $1$e-$4$ & $0$ & $2$ & - & -\\
        GCN\_LC & Reddit & $1$e-$7$ & $0.001$ & $0.5$ & $2$ & - & -\\
        & arxiv & $1$e-$4$ & $0.001$ & $0$ & $2$ & - & -\\
        & papers100M & $1$e-$6$ & $0.001$ & $0$ & $2$ & - & - \\ \hline
        & Flickr & $1$e-$4$ & $1$e-$4$ & $0$ & $3$ & concat & - \\
        JKNet\_LC & Reddit & $1$e-$6$ & $1$e-$4$ & $0.5$ & $3$ & concat & -\\
        & arxiv & $1$e-$6$ & $1$e-$4$ & $0.5$ & $3$ & concat & -\\
        & papers100M & $1$e-$6$ & $0.001$ & $0$ & $3$ & concat & -\\ \hline
        & Flickr & $0$ & $1$e-$4$ & \begin{tabular}{c}MLP : $0.5$\\ Propagation : $0$ \end{tabular} & \begin{tabular}{c}MLP: $4$\\ Propagation : $5$ \end{tabular} & - & $0.1$ \\
        GPRGNN\_LC & Reddit & $0$ & $1$e-$4$ & \begin{tabular}{c}MLP : $0.5$\\ Propagation : $0$ \end{tabular} & \begin{tabular}{c}MLP: $4$\\ Propagation : $5$ \end{tabular} & - & $0.5$\\
        & arxiv & $1$e-$8$ & $0.001$ & \begin{tabular}{c}MLP : $0.5$\\ Propagation : $0.5$ \end{tabular} & \begin{tabular}{c}MLP: $4$\\ Propagation : $5$ \end{tabular} & - & $0.9$\\
        & papers100M & $1$e-$7$ & $0.001$ & \begin{tabular}{c}MLP : $0.5$\\ Propagation : $0$ \end{tabular} & \begin{tabular}{c}MLP: $4$\\ Propagation : $5$ \end{tabular} & - & $1$\\ \bottomrule
    \end{tabular}
    }
\label{tb:best_param}
\end{table*}
\end{document}